\documentclass{article}

\usepackage{arxiv}

\usepackage[utf8]{inputenc}
\usepackage{graphicx}
\usepackage[noadjust]{cite} 
\usepackage{amsmath}
\usepackage{hyperref}
\hypersetup{colorlinks=true, citecolor=blue}
\usepackage{tikz}
\usetikzlibrary{positioning}
\usepackage{tcolorbox}
\usepackage{nicematrix, booktabs,pgfplots}
\usepackage{tikz-network}
\usepackage{geometry}
\usepackage{booktabs}
\usepackage{tabularx}
\usepackage{scalerel} 
\usepackage{pgfplots}
\usepgfplotslibrary{groupplots}
\usepackage{booktabs}
\usepackage{siunitx}
\usepackage{amsmath,amsfonts}
\usepackage{algorithm}
\usepackage{algpseudocode}
\usepackage{array}
\usepackage[caption=false,font=normalsize,labelfont=sf,textfont=sf]{subfig}
\usepackage{textcomp}
\usepackage{stfloats}
\usepackage{url}
\usepackage{verbatim}
\usepackage{graphicx}
\usepackage{cite}
\usepackage{tipa}
\usepackage{eurosym}
\usepackage{multirow}

\usepackage{natbib}

\DeclareMathOperator*{\concat}{\scalerel*{\Vert}{\sum}}

\usetikzlibrary{
	arrows.meta,         
	matrix,              
}

\tikzset{
	rows/.style 2 args={
		/utils/temp/.style={row ##1/.append style={nodes={#2}}},
		/utils/temp/.list={#1}},
	columns/.style 2 args={
		/utils/temp/.style={column ##1/.append style={nodes={#2}}},
		/utils/temp/.list={#1}}}
\makeatletter
\tikzset{
	left delimiter/.style 2 args={append after command={
			\tikz@delimiter{south east}{south west}
			{every delimiter,every left delimiter,#2}{south}{north}{#1}{.}{\pgf@y}}}}
\makeatother

\title{Inductive inference of gradient-boosted decision trees on graphs for insurance fraud detection}

\author{
	Félix Vandervorst \\
	Data Office, Allianz Benelux\\
	KU Leuven\\
	University of Antwerp-imec
	\And
	Bruno Deprez\thanks{Correspondening author: \href{mailto:bruno.deprez@kuleuven.be}{\texttt{bruno.deprez@kuleuven.be}}} \\
	KU Leuven\\
	University of Antwerp-imec
	\And
	Wouter Verbeke\\
	KU Leuven
	\And
	Tim Verdonck\\
	University of Antwerp-imec\\
	KU Leuven 
}

\begin{document}
\maketitle

\begin{abstract}
	Graph-based methods are becoming increasingly popular in machine learning due to their ability to model complex data and relations. 
	Insurance fraud is a prime use case, since fraudulent claims are often the result of organised criminals that stage accidents or the same persons filing erroneous claims on multiple policies. 
	One challenge is that graph-based approaches struggle to find meaningful representations of the data because of the high class imbalance present in fraud data. 
	In addition, insurance graphs are heterogeneous and dynamic, given the changing relations among people, companies and policies. 
	As a result, gradient-boosted tree approaches on tabular data still dominate the field. 
	Therefore, we present a novel inductive graph gradient boosting machine (G-GBM) for supervised learning on heterogeneous and dynamic graphs. 
	G-GBM combines the class-imbalance robustness of gradient boosting with heterogeneous graph information encoded through interpretable path-level feature concatenations, while preserving access to the original tabular feature space. 
	In addition, the explicit representation of neighbourhood information enables transparent SHAP-based explanations at the metapath and feature level.
	We demonstrate G-GBM for insurance fraud detection on an open-source and a real-world, proprietary dataset, and find that G-GBM performs on par or better than the state-of-the-art. 
	The associated insurance fraud dataset is publicly released to facilitate reproducibility. 
\end{abstract}


\section{Introduction}
Fraud costs insurance companies billions every year. 
This cost is eventually integrated in the premium and hence passed on to the clients. 
In the United States, the Coalition Against Insurance Fraud estimates that fraud costs more than \$ 300 billion per year \citep{CAIF21}. 
The Federal Bureau of Investigation (FBI) estimates that non-health insurance fraud is worth more than \$40 billion annually \citep{FBI21}. 
In the EU, the European (re)insurance federation estimates that insurance fraud amounts to \euro{} 13 billion annually and that the total of all cases of fraud---both detected and undetected---amounts to 10\% of the total claims expenditure in Europe \citep{insuranceeu21}. 
In the United Kingdom, the Association of British Insurers reported that in 2021 alone, insurers detected 89,000 dishonest insurance claims valued at £ 1.1 billion \citep{ABIFraud21}. 

An effective strategy to detect and prevent fraud is critical to reduce costs and gain a competitive advantage with a lower insurance premium.
Supervised and unsupervised learning methods have been the most popular approaches to (insurance) fraud since the 1990s \citep{bolton2002statistical}.
These approaches aim to exploit historical patterns to predict fraudulent cases for further inspection.
These approaches assume that historical insurance claims are \textit{independent and identically distributed} (i.i.d.). 
However, this is not suited for effective identification of fraud hubs, which are of interest, not only for insurance companies, but also for law enforcement agencies, since organised insurance fraud is used to launder money and/or fund further criminal activities.

Graph-based methods have offered new avenues for detecting unexplored patterns and have become a popular method to insurance fraud detection~\citep{oskarsdottir2020social,campoEngine,deprez2024insurance}. 
Unlike traditional methods, graph-based methods explicitly model the relationships between individuals, making them appropriate for detecting organized insurance fraud, where multiple parties collaborate to commit fraud \citep{sadowski2014fraud}. 

The complex interaction structure between different parties results in a heterogeneous graph with diverse node and edge types. 
This heterogeneity is intrinsic to insurance graphs, since people and companies, i.e., the clients, interact with the insurer through different policies over their lifetime. 
Previous graph analytics studies often capture this heterogeneity by constructing graph embeddings based on metapaths in the graph~\citep{10.1145/2124295.2124373,wang2019heterogeneous,10.1145/3366423.3380297,cao2018collective,chen2019infdetect} or apply graph neural networks~(GNNs)~\citep{deprez2024insurance}. 

There are three main challenges when applying graph-based deep learning methods to insurance fraud detection.
The first drawback is that graph structures evolve over time, changing a node's connections, features and even labels. 
There is thus a need for inductive methods that update the predicted probability for a node to be fraudulent whenever new information becomes available. 
Traditional embedding approaches based on static random walks, such as node2vec~\citep{grover2016node2vec} or metapath2vec~\citep{dong2017metapath2vec}, are less suited for settings where graph structures evolve over time and predictions need to be updated inductively.
Most heterogeneous GNNs (HGNNs) incorporate metapaths as a fundamental part of their architecture~\citep{10.1145/3447548.3467350, 7536145}. Since the metapaths are only applied for (iterative) feature aggregation purposes, these can also be used in an inductive setting. 
However, the second drawback is that fraud detection involves mostly tabular data with a high class imbalance. 
In tabular settings with strong class imbalance and heterogeneous feature types, tree-based methods often remain highly competitive compared to deep learning approaches~\citep{Grinsztajn_tree_vs_dl}, especially compared to gradient boosting methods like LightGBM~\citep{ke2017lightgbm}. 

A final challenge arises from the structure of the data itself. 
Client information often contains many high-cardinality categorical features. 
These must be converted to continuous values in the pre-processing steps so they can be used by HGNNs. 
Even then, the neighbourhood feature aggregation of most HGNNs may be less suitable for high-cardinality categorical variables. 
For example, aggregating categorical variables such as car brands through averaging operations is generally not meaningful.
Tree-based methods are better suited to deal with these categorical features. 

To summarise, insurance fraud detection offers a unique and challenging use case, since (1) graph information is invaluable, (2) the graph structure is inherently heterogeneous and (3) dynamic, and (4) only few cases are ever investigated resulting in a high class imbalance. 
In this work, we contribute to the state-of-the-art in supervised learning on complex graph structures as applied for, e.g., predict insurance fraud risk. 
We present a novel method called graph-gradient boosted machine (G-GBM), which combines the detection power of gradient boosting with the complex data representation of heterogeneous graphs. 

G-GBM inherits the strengths of LightGBM — handling class imbalance, non-linear relationships, and categorical features — while extending it to heterogeneous graph topologies through probability-weighted metapath featurization. 
Crucially, neighbourhood information is encoded as explicit concatenations of node and edge features along each path, rather than aggregated into opaque embeddings. 
This means every feature driving a prediction can be traced back to a specific node or edge in the graph via SHAP values, providing the audit trail that is a critical requirement in the regulated insurance industry. 
This design also guarantees that G-GBM performs at least as well as a tabular LightGBM baseline, and avoids the over-smoothing that plagues heterogeneous GNNs when additional message-passing layers are added.

We compare the performance of G-GBM to feature-based classification without graph information and to different HGNN methods on heterogeneous graphs for insurance fraud detection.
These experiments demonstrate that G-GBM achieves competitive predictive performance. 
Additionally, given that  G-GBM utilises tree-based learners, we illustrate how the popular SHAP-based explanations can naturally be derived in classic supervised learning problems. 

Hence, the main contributions of our work are as follows:
\begin{itemize}
    \item \textbf{Theoretical:} G-GBM provides an inductive framework for supervised learning on heterogeneous graphs with a formal Pareto dominance guarantee over its tabular counterpart, as the graph structure can only help performance.
    \item \textbf{Methodological:} The probability-weighted metapath featurization with adapted Gini impurity offers a theoretically grounded bridge between random-walk path semantics and gradient boosting, addressing the open problem of tree-based learning on heterogeneous, dynamic graphs.
    \item \textbf{Empirical: }On real-world insurance fraud data, G-GBM achieves strong predictive performance while preserving the practical advantages of GBMs over GNNs, including treatment of categorical features, handling of missing data, and path-level SHAP audit trails that link each prediction to specific nodes and edges in the graph.
    \item \textbf{Reproducibility \& benchmarking: } We release both the full implementation on GitHub ({\url{https://github.com/VerbekeLab/GBDT_Graphs}}) and the anonymized real-world Belgian insurance fraud graph ({\url{https://www.kaggle.com/dsv/16002794}} \citep{f_lix_vandervorst_bruno_deprez_tim_verdonck_wouter_verbeke_2026}) as open resources, providing the community with a rare proprietary-grade heterogeneous graph benchmark for fraud detection research. 
\end{itemize}

\section{Related work}
\label{sec:related_work}

\subsection{Learning on tabular data.}
Machine learning methods for fraud detection (including insurance) are essentially dominated by two approaches \citep{bolton2002statistical,abdallah2016fraud}: supervised and unsupervised learning (or a combination thereof). 

Insurance fraud detection poses several distinctive challenges. 
First, only a limited number of fraud cases are labelled. This results both from fraud being a rare event and by the fact that the majority of claims are never investigated due to resource constraints. 
Second, fraudsters continuously change tactics, making fraud detection model trained on historical data rapidly obsolete~\citep{van2017gotcha,baesens2015fraud}.

Therefore, unsupervised learning has been widely applied for fraud detection, as it does not rely on (historical) labels for training. 
Unsupervised learning mostly relies on anomaly detection, as fraudulent claims are assumed to be exaggerated or unrealistic, and hence differ substantially from normal claims~\citep{https://doi.org/10.1111/jori.12427,STRIPLING201813,NIAN201658,VANDERVORST2022113798}.

Although unsupervised learning seems the obvious choice when dealing with sparsely labelled data, fraud is also imperceptibly concealed~\citep{van2017gotcha}.
As a result, not all fraud cases are anomalies and not all anomalies are fraud. 
Previous research on insurance fraud as well as other types of financial fraud has shown that unsupervised learning often underperforms compared to supervised learning~\citep{https://doi.org/10.1111/jori.12427, 10.1145/3383455.3422549,deprez2024networkevaluation}.

Supervised learning in the context of fraud is typically addressed as a binary classification problem. 
Historical information on claims and fraud labels are assumed to be \textit{independent} realizations of random variables. 
One generally seeks to determine the posterior distribution $\eta(X) := P(Y=1{\vert}X)$ to estimate the probability of fraud based on feature values $x$ only. 
Although it is well known that there is no one method that performs better in every single case, some methods seem generally more powerful and are more commonly used in practice than others. 
Popular state-of-the-art methods in machine learning are gradient-boosted tree methods, such as LightGBM \citep{ke2017lightgbm}, XGBoost \citep{chen2016xgboost}, and CatBoost \citep{prokhorenkova2018catboost}. 
These models often remain highly competitive compared to neural networks on tabular datasets \citep{shwartz2022tabular,Grinsztajn_tree_vs_dl}.

\subsection{Learning with Graph Features}
More recently, graph learning methods have gained traction in the field of fraud detection. 
Graphs can be used to represent the relationships between individuals or entities. 
Relational patterns may become visible when comparing entities to related ones in the graph. 
The application of graph features has been essential for enhancing methods across fields, such as social security fraud detection~\citep{van2017gotcha}, churn prediction~\citep{VERBEKE2014431}, anti-money laundering~\citep{deprez2024networkevaluation} and uncovering malicious online ratings~\citep{SUN2021106895, SUN2024121236}.

Graph information has been shown to be particularly useful for detecting organized insurance fraud \citep{sadowski2014fraud,deprez2024insurance}, as well. 
A common approach to learn from graphs involves enriching the classical supervised learning problem with new features that are generated from the graph structure to supplement the  ``intrinsic'' (node) features~\citep{oskarsdottir2020social,deprez2024insurance}. 
These new graph features can be based on simple graph statistics (e.g., degree and betweenness), so-called guilt-by-association features (e.g., PageRank~\citep{page1999pagerank}), random walk-based features (e.g., DeepWalk~\citep{perozzi2014deepwalk} and Node2Vec~\citep{grover2016node2vec}) or neural network-based features (e.g., graph convolutional networks~\citep{kipf2016semi} and GraphSAGE~\citep{hamilton2017inductive}). 

In the context of (insurance) fraud, ``guilt by association'' has been proven to be particularly effective \citep{vsubelj2011expert,oskarsdottir2020social,menon2018information,van2017gotcha,deprez2024insurance}.
\citet{calderoni2020robust} examined the structure of a criminal network and demonstrated the effectiveness of the Personalized PageRank algorithm. 
They emphasize the pivotal role of topology selection in graph construction, noting that the choice of modelled relations significantly impacts the results. 
This finding underscores the critical importance of graph design parameters in analysing criminal networks. 

In the context of detecting motor insurance fraud, \citet{vsubelj2011expert} presented different possible topologies and highlighted the contrast between utilizing expressive topologies versus simpler representations. 
In social security fraud, \citet{van2017gotcha} introduce methods to summarize graph characteristics for supervised downstream classification. 
\citet{cao2018collective,chen2019infdetect} also used ``intrinsic'' node features supplemented with metapath information (including proximity to other fraud labels) for e-commerce fraud. 
However, the choice of graph-based features is arbitrary and depends on the application. 
Recently, graph neural networks have increasingly been used to generalizing the discovery of graph-based patterns \citep{deprez2024insurance}.

\subsection{Deep Learning on Graphs}
A recent trend in graph analysis concerns graph embedding, which aims to condense information from a graph's neighbourhood into a single feature vector.
These methods map node and neighbourhood information into a lower-dimensional Euclidean representation, aiming to conserve as much topological information using as small a dimension as possible.
This trend encompasses three distinct methodologies: (i) methods based on matrix factorization, (ii) techniques based on random walks, such as DeepWalk~\citep{perozzi2014deepwalk} and Node2Vec~\citep{grover2016node2vec}, and (iii) more recent approaches utilizing graph neural networks (GNNs), such as GraphSAGE~\citep{hamilton2017inductive}.

GNNs take advantage of more information by incorporating the node features directly. 
They iteratively aggregate neighbourhood information, based on message passing. 
The final GNN layer is task specific, e.g., to make fraud predictions.
This allows to incorporate information on node features, graph topology and the downstream task into the graph embedding.

When learning on heterogeneous graphs, the message passing is split and learned according to the relation type~\citep{10.1007/978-3-319-93417-4_38}. 
Since not all relation types are as meaningful, many of the HGNNs are based on metapaths that define which relations are of interest. 
\citet{wang2019heterogeneous} introduced the graph attention network (HAN), which learns attention weights both at the node and metapath level.
The heterogeneous graph transformer (HGT)~\citep{10.1145/3366423.3380027} incorporates transformer-based message passing across heterogeneous relations and is designed to efficiently scale to large heterogeneous graphs.

In the context of fraud detection, \citet{van2022inductive, van2023catchm} used graph embedding for credit card fraud prediction, particularly HinSAGE, a heterogeneous graph extension of the popular GraphSAGE~\citep{hamilton2017representation} method implemented in StellarGraph~\citep{StellarGraph}. 
This work was further extended by \citet{10478631} to be better suited to deal with skewed class distributions. 
\citet{deprez2024insurance} compared HinSAGE to other graph embedding methods on insurance fraud data.

Beyond applications in fraud detection, there is considerable ongoing research in graph-based machine learning that focuses on more complex and expressive forms of graphs, including those that are heterogeneous and time dependent. 
As an extensive discussion on graph neural networks (GNNs) is beyond the scope of this paper, we refer to \citet{shi2022heterogeneous} for a recent comprehensive overview of HGNNs.

As noted in \citet{hamilton2017representation}, there is much work to do to improve the theoretical framework to guide future graph-based methods. 
In particular, in the field of heterogeneous graph neural networks, a theoretical framework is lacking \citep{shi2022heterogeneous}. 
Several open challenges remain regarding the reliability of graph learning and its interpretability, which is particularly important for fraud applications, and to address methodological concerns of dependence between training and testing on overlapping graphs, a challenge that remains insufficiently explored in parts of the graph learning literature.
\section{Preliminaries}
\label{sec:preliminaries}
This section introduces the main graph-learning concepts used throughout the proposed G-GBM framework.
To improve the mathematical clarity and accessibility of our proposed G-GBM framework, we summarize the main notations in Table~\ref{tab:notation}.

\begin{table*}[]
    \centering
    \caption{Summary of key notations and definitions.}
    \label{tab:notation}
    \begin{tabular}{lll}
    \toprule
    Symbol & Name/Definition & Section/Equation \\
    \midrule
    $\mathcal{G}$ & Heterogeneous Information Network (HIN) & Section~\ref{subsec:HIN}\\
    $\mathcal{V, E}$ & Sets of nodes and edges, respectively & Section~\ref{subsec:HIN}\\
    $\phi, \psi$ & Node and edge type mapping functions, respectively & Section~\ref{subsec:HIN}\\
    $\chi, \xi$ & Node and edge feature mapping functions, respectively & Section~\ref{subsec:HIN}\\
    $N_v^k$ & Nodes at exactly distance $k$ from node $v$ & Section~\ref{subsec:egonets}\\
    $\mathcal{G}_v^n$ & $n$-hop neighbourhood of node $v$ & Equations~\eqref{eq: def V}-\eqref{eq: def E}\\
    $\mathcal{P}_v^n$ & Set of paths starting at node $v$ of length at most $n$ in $\mathcal{G}_v^n$ & Equation~\eqref{eq: def simple path}\\
    $P_{v,w}^n$ & A path from $v$ to $w$ in $\mathcal{G}_v^n$ & Section~\ref{subsec:egonets}\\
    $\mathbb{P}$ & Probability measure for random walker taking a path & Equation~\eqref{eq:weight_metapath} \\
    $h(P_{v,w}^n)$ & $n$-length feature function of path $P_{v,w}^n$ & Equation~\eqref{eq:feature_concat} \\
    $\hat{\eta}(\mathcal{G}^n_v)$ & The G-GBM fraud probability estimator & Equation~\eqref{eqn:weighted_loss} \\
    \bottomrule
\end{tabular}
\end{table*}

\subsection{Heterogeneous Information Networks}
\label{subsec:HIN}
A \textbf{heterogeneous information network (HIN)} is a graph containing multiple types of nodes and/or edges, denoted as \( \mathcal{G}(\mathcal{V}, \mathcal{E}, \phi, \psi, \chi, \xi) \) with
\begin{itemize}
    \item \( \mathcal{V} \) is the set of nodes.
    \item \( \mathcal{E} \) is the set of edges; \( \mathcal{E} \subset \mathcal{V}\times\mathcal{V} \).
    \item \( \phi: \mathcal{V} \rightarrow \mathcal{T} \) is the node type mapping function, with \( \phi(v) \in \mathcal{T}, \forall v \in \mathcal{V} \), indicating the unique node type for node every \(v\).
    \item \( \psi: \mathcal{E} \rightarrow \mathcal{R} \) is the edge relation type mapping function, with \( \psi(e) \in \mathcal{R}, \forall e \in \mathcal{E} \), indicating the unique relation type for every edge \(e\).
    \item \( \chi: \mathcal{V} \rightarrow \bigcup_{\tau \in \mathcal{T}} \mathbb{R}^{d_\tau} \) is the node feature mapping function, where each node \( v \) is assigned to a feature vector in space \( \mathbb{R}^{d_\tau} \). Dimension \( d_\tau \) corresponds to node type \( \tau = \phi(v) \), allowing the dimension to vary depending on the node type.
    \item \( \xi: \mathcal{E} \rightarrow \bigcup_{\rho \in \mathcal{R}} \mathbb{R}^{f_\rho} \) is the edge feature mapping function, where each edge \( e \) is mapped to a feature vector in a space \( \mathbb{R}^{f_\rho} \). Dimension \( f_\rho \) corresponds to edge type \( \rho = \psi(e) \), allowing the dimension to vary depending on the edge type.
\end{itemize}
A graph having only one type of node and edge is called a \textbf{homogeneous graph}, i.e., $|\mathcal{T}| = 1$ and $|\mathcal{R}|=1$ where $|\mathcal{T}|$ and $|\mathcal{R}|$ denote the numbers of node and edge types, respectively.

\subsection{Local graph information: $n$-hop neighbourhood and metapaths}
\label{subsec:egonets}
\begin{figure}
    \centering
    \scalebox{0.7}{\begin{tikzpicture}[node distance=0.9cm]
  \tikzstyle{Ego}=[circle,draw,fill=orange!30]
  \tikzstyle{Admin}=[circle,draw,fill=green!60]
  \tikzstyle{Company}=[circle,draw,fill=orange!60]

  \node[Ego] (ego) {$0$};

  \node[Admin] (n1) [above left=of ego] {1};
  \node[Company] (n2) [above=of ego] {2};
  \node[Company] (n3) [above right=of ego] {3};
  \node[Company] (n4) [right=of ego] {4};
  \node[Company] (n5) [below=of ego] {5};
  
  \draw (ego) -- (n1);
  \draw (ego) -- (n2);
  \draw (ego) -- (n3);
  \draw (ego) -- (n4);
  \draw (ego) -- (n5);

  \node[Company] (n1a) [above left=of n1] {6};
  \node[Admin] (n2a) [above=of n2] {7};
  \node[Admin] (n4a) [right=of n4] {8};
  \node[Company] (n5a) [below right=2.2cm] {9};

  \draw (n1) -- (n1a);
  \draw (n2) -- (n2a);
  \draw (n4) -- (n4a);
  \draw (n4) -- (n5a);
  \draw (n5) -- (n5a);

    \draw[dotted, thick] (ego) circle (2.2cm) node[below left = 0.6cm] {$G_v^1$};
    \draw[dotted, thick] (ego) circle (4cm) node[below left = 2.3cm and 1.4cm] {$G_v^2$};

\end{tikzpicture}}
    \caption{A $2$-hop ego-net $G^n_v$ centred on node $v$ with two node types: companies (orange) and administrators (green). The simple path set $P(v)^n$ (edges labels are omitted for visual clarity) is: $\{(v_0, v_1, v_6), (v_0, v_2,v_7), (v_0, v_3), (v_0,v_4,v_8),$
    $(v_0, v_4, v_9), (v_0,v_5,v_9)\}$}
    \label{fig:exampleEgonet}
\end{figure}
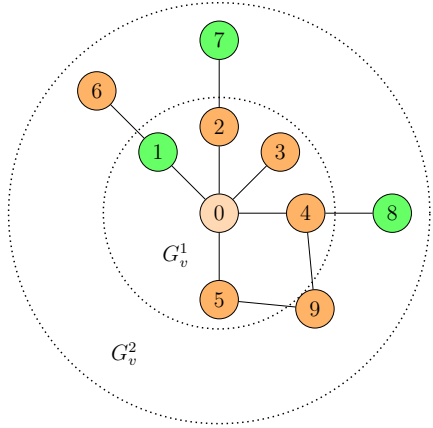

Let \( N_v^k \) be the set of nodes that are exactly \( k \) steps away from \( v \) for a node $v$ in graph $\mathcal{G}$.
The \textbf{$\boldsymbol{n}$-hop ego-net} \( \mathcal{G}_v^n \) of the vertex $v$, as presented in Figure~\ref{fig:exampleEgonet}, is the induced sub-graph of \( \mathcal{G}(\mathcal{V}, \mathcal{E}) \) with the vertex set \(\mathcal{V}_v^n\subseteq\mathcal{V}\) and the edge set \( \mathcal{E}_v^n \subseteq \mathcal{E} \) defined as follows:
\begin{eqnarray}
\mathcal{V}_v^n &=& \{ v \} \cup \bigcup_{k=1}^{n} N_v^k, \label{eq: def V} \\
\mathcal{E}_v^n &= &\{ (u, w) \in \mathcal{E} | u, w \in \mathcal{V}_v^n \}. \label{eq: def E}
\end{eqnarray} 

A \textbf{simple path} of length $k$ in graph $\mathcal{G}$ from node \( v \) to node \( v_k \) is a sequence of nodes and edges \( (v_0=v, e_1, v_1, e_2, ..., e_k, v_k) \) such that

\begin{itemize}
    \item For each \( i \) (1 $\leq$ \( i \) $\leq$ \( k \)), \( e_i = (v_{i-1}, v_i) \in \mathcal{E} \), each pair of consecutive nodes in the path is connected by an edge in the graph.
    \item All nodes in the sequence are distinct: for any 0 $\leq$ \( i, j \) $\leq$ \( k \), \( i \) $\neq$ \( j \), \( v_i \) $\neq$ \( v_j \).
\end{itemize} 
We define the \textbf{set of simple paths} in the $n$-hop ego-net $\mathcal{G}^n_v$ as: \begin{equation}
    \mathcal{P}_v^n= \{ P \mid P\text{ is a simple path of length at most } n\text{ in }\mathcal{G}^n_v\} \label{eq: def simple path}
\end{equation}

The types of nodes and edges may vary in a simple path, reflecting the heterogeneous nature of the graph. 
Each node \( v_i \) can be of any type \( \phi(v_i) \in \mathcal{T} \) and each edge \( e_i \) can be of any type \( \psi(e_i) \in \mathcal{R} \).
Alternatively, it is common to use a \textbf{metapath}, which is defined as a specific sequence of node types, $T_i\in \mathcal{T}$, and edge types, $R_i\in\mathcal{R}$ \citep{dong2017metapath2vec}:
\[\mathcal{P}:T_1 \xrightarrow{R_1} T_2 \xrightarrow{R_2} \ldots T_t \xrightarrow{R_t} T_{t+1}\ldots \xrightarrow{R_{l-1}}V_l. \]

The definition of metapaths has the advantage of explicitly representing only possible paths, which can facilitate computations when this information is available.
Suppose in insurance, we have a graph with nodes representing policy holders (P), claims (C), addresses (A) and vehicles (V). 
The edges are defined using the following relation: $ P \xrightarrow{\text{files}} C$, $C \xrightarrow{\text{involves}} V$, $P \xrightarrow{\text{lives\_at}} A$. 
For fraud ring detection, it might be most interesting to construct graph learning models using the metapaths such as VCV, i.e., looking at vehicles that appear in the same claims. 
More complex relational structures can also be represented.
The metapath CPAPC represents claims filed by people living at the same address. 

\section{Methodology}
In this section, we discuss elements from graph learning to introduce our novel method: G-GBM. We show how node and edge features are combined using metapaths. Afterwards, we present our G-GBM method, which uses probability-weighted featurised paths to exploit the information from the node and edge features. A formal description of G-GBM in pseudo-code is provided in Algorithm~\ref{alg:g-gbm}.

\begin{algorithm}
    \caption{ G-GBM }\label{alg:g-gbm}
    \textbf{Input}: Graph $\mathcal{G}(V,E, \phi, \psi, \chi, \xi)$, metapath schema $\mathcal{P}$\\
    \textbf{Output}: Predictions $\hat{\eta}\left(\mathcal{G}^n_v\right)$ for node $v$ based on its $n$-hop ego-net.
    \begin{algorithmic}[1]
        \State Sample all metapaths according to $\mathcal{P}$ starting at $v$
        
        \For {metapaths $p$}
        	\State Calculate weights $\mathbb{P}(P_{v,p}^n)$ according to Equation~\eqref{eq:weight_metapath}
            \State Calculate features $x_{v,p}^n$ according to Equations~\eqref{eq:feature_concat}-\eqref{eq:feature_function_edge}
        \EndFor 
        \State Train gradient boosting forest $\hat{\eta}_v^p(x_{v,p}^n)$
        \State $\hat{\eta}(\mathcal{G}^n_v) = \sum_p \mathbb{P}(P_{v,p}^n)\hat{\eta}_v^p(x_{v,p}^n)$
    \end{algorithmic}
\end{algorithm}

\subsection{Ego-net paths to supervised learning}
In a first step, neighbourhood information is aggregated based on sampled metapaths in the $n$-hop neighbourhood of each node. 
Each path is assigned a relative weight, used as case weights in the down-stream decision tree algorithm. 
Let us denote the probability of a random walker taking a path from $v$ to $p$ by $\mathbb{P}(P_{v,w}^n)$. By construction, $\sum_w \mathbb{P}(P_{v,w}^n) = 1$.
Then, $\mathbb{P}((v_0,e_1,v_1, ... e_k,v_k))$ is defined as:
\begin{equation}
    \mathbb{P}\left((v_0,e_1,v_1, ... e_k,v_k)\right) := \prod_{i=1}^{k} p\left(v_{i-1}, e_i, v_i| \mathcal{G}^{(i-1)}\right),
    \label{eq:weight_metapath}
\end{equation}
where $\mathcal{G}^{(i)} := \mathcal{G}^{(i-1)} \backslash \{v_{i-1}\}$ for $i > 0$ is the subgraph of $\mathcal{G}$ excluding the entries visited at step $(i)$ in the simple path and $\mathcal{G}^{(0)} = \mathcal{G}$.

The representation of node and edge attributes along a given simple path is defined as follows.
We define the $n$-length featurisation function $h: P \to \{R, \emptyset\}^{n \times (\sum_{i \in \mathcal{T}} d_i +  \sum_{j \in \mathcal{R}} d_j)}$ of path $P_{v,p}^n$ as the concatenation of node and edge features as encountered on the path $P_{v,p}^n$:

\begin{eqnarray}
    h(P_{v,p}^n) &:= & \left(\concat_{i \in \mathcal{T}} \chi_i(v_0)\right) \concat \nonumber \\ 
    & &\left[ \concat_{k=1}^n \left(\left(\concat_{j \in \mathcal{R}} \xi_j(\{v_{k-1},v_{k}\})\right) 
    \concat \left(\concat_{i \in \mathcal{T}} \chi_i(v_k)\right)\right) \right] \nonumber \\ 
    &:= &x_{v,p}^n \label{eq:feature_concat}
\end{eqnarray}
where $(v_0, ..., v_{n'})$ is the order sequence of vertices in $P_{v,p}^n$. 
We note that $n' \leq n$ allows for early stopping, in case the path arrives at a dead end. 
The functions $\chi_i$, $\xi_j$ assign the feature values:
\begin{eqnarray}
    \chi_i(v_k) &=& 
    \begin{cases}
        \chi(v_k) & \text{if }\phi(v_k)=\mathcal{T}^i \text{ and } k<n'\\
        \emptyset^{d_i} & \text{otherwise}
    \end{cases}, \label{eq:feature_function_node}
\\
    \xi_j(\{ v_{k-1}, v_k \}) &=& \begin{cases}
        \xi(\{ v_{k-1}, v_k \}) & \text{if }\psi(\{ v_{k-1}, v_k \}) = \mathcal{R}^j \text{ and } k<n' \\
        \emptyset^{d_j} & \text{otherwise}
    \end{cases},\label{eq:feature_function_edge}
\end{eqnarray}
where $k<n'$ expresses that the path is shorter than the maximal length. 

Note that the above definition can be simplified for specific graph topologies. For instance, an HIN with multiple node types and features but only one type of edge, with no edge features, simplifies to
\begin{equation}
     h(P_{v,p}^n) := \concat_{k=0}^n \Big(\concat_{i \in \mathcal{T}} \chi_i(v_k)\Big),
\end{equation}
which is the concatenation of node features encountered on the metapath.

Ultimately, we are interested in classifying a node as fraudulent or non-fraudulent, based on these featurised path vectors. 
The loss function of the problem is written as $\mathcal{L}(y_v,\hat{y}_v)$ for any node $v$.
Our proposed method is an estimator $\eta(\mathcal{G}^n_v)$ based on its metapath predictions weighted by their random walk probabilities, as follows:
\begin{equation}
    \hat{\eta}(\mathcal{G}^n_v) = \sum_p \mathbb{P}(P_{v,p}^n)\hat{\eta}_v^p(x_{v,p}^n)
    \label{eqn:weighted_loss}
\end{equation}
where $\hat{\eta}^p :=\hat{\mathbb{P}}(y_{v}=1 | x_{v,p}^n)$.
In other words, the probability of fraud of a given node $v$ with respect to its local neighbourhood $\mathcal{G}^n_v$ is the sum of their estimated probability of path fraud, weighted by the probability of a random walker taking this path. 
Equation~\eqref{eqn:weighted_loss} can be interpreted as a weighted loss function.

We also note the case of either a $0$-hop ego-net or a fully disconnected graph (i.e., with no edges). For a supervised learning task where the head node is of one specified type, our framework reduces to the classical supervised learning setting. Indeed, each node presents only one simple path (itself), so $|p|=1$, $\mathbb{P}(P_{v,p}^n)=1$, $x^n_{v,p}$ refers to samples $x_v$, and our estimator of Equation~\eqref{eqn:weighted_loss} becomes the classic supervised learning estimator $\hat{\mathbb{P}}(y_{v}=1 | x_{v})$.

The fraud probabilities of G-GBM are determined using an adaptation of gradient boosted decision trees, with case weights to estimate $\hat{\eta}_v$.
In decision tree algorithms, the Gini impurity measures how often a randomly chosen element from the set would be incorrectly labeled if it were randomly labeled according to the distribution of labels in the subset. 
It is used as a split criterion in constructing a decision tree.
For a set \( S \) containing samples of various classes, the Gini impurity is calculated as:
\begin{equation}
\text{Gini}(S) = 1 - \sum_{i=1}^{|\mathcal{Y}|} (\Delta_i)^2,
\label{eq:gini}
\end{equation}
where $|\mathcal{Y}|$ is the number of classes and \( \Delta_i \) is the proportion of samples in \( S \) that belong to class \( i \).

In the context of sample weights, the proportion \( \Delta_i \) is computed by summing the weights (for our method, the probability of the metapaths) of the samples in each class instead. 
The weighted Gini impurity for a set \( S \) is:
\begin{equation}
\text{Gini}_{\text{w}}(S) = 1 - \sum_{i=1}^{|\mathcal{Y}|} \left(\frac{\sum_{x_{v,p}^n \in S_i} \mathbb{P}(P_{v,p}^n)}{\sum_{x_{v,p}^n \in S} \mathbb{P}(P_{v,p}^n)}\right)^2,
\label{eq:gini_weight}
\end{equation}
where \( S_i \) is the subset of \( S \) that belongs to class \( i \), and \( \mathbb{P}(P_{v,p}^n) \) is the weight of the sample \( x_{v,p}^n \).

When considering a split of \( S \) into two subsets \( S_L \) and \( S_R \), the weighted Gini impurity for the split is a weighted average of the impurities of the two subsets:
\begin{eqnarray}
\text{Gini}_{\text{split}}(S_L,S_R) & = & \frac{\sum_{x_{v,p}^n \in S_L} \mathbb{P}(P_{v,p}^n)}{\sum_{x_{v,p}^n \in S} \mathbb{P}(P_{v,p}^n)} \text{Gini}_{\text{w}}(S_L) \nonumber \\
& + &\frac{\sum_{x_{v,p}^n \in S_R} \mathbb{P}(P_{v,p}^n)}{\sum_{x_{v,p}^n \in S} \mathbb{P}(P_{v,p}^n)} \text{Gini}_{\text{w}}(S_R)
\end{eqnarray}

The algorithm will choose the split that results in the lowest weighted Gini impurity, which implies the greatest increase in the ``purity'' of the resulting subsets, taking into account the case weights.

Furthermore, our method can generate missing values (denoted by $\emptyset$ in the above). The Gini criterion is adapted as follows: consider $S_L' := (S_L \cup S_{\empty})$ and $S_R' := (S_R \cup S_{\empty})$.
\begin{equation}
    \text{Gini}_{\text{split}}' = arg max\left(\text{Gini}_{\text{split}}(S_L',S_R), \text{Gini}_{\text{split}}(S_L,S_R')\right)
\end{equation}

In LightGBM~\citep{ke2017lightgbm}, XGBoost~\citep{chen2016xgboost}, CatBoost~\citep{prokhorenkova2018catboost} and other gradient boosting frameworks that support case weights, this concept is applied during the construction of trees. 
Since $\sum_p \mathbb{P}(P_{v,p}^n) = 1$ and all the featurised paths of node $v$ belong to the same class as its head node, Equation~\eqref{eq:gini_weight} reduces to Equation~\eqref{eq:gini}. 

In summary, our G-GBM method presents multiple key characteristics that make it ideally suited for insurance fraud detection on heterogeneous graph data. 
First, the backbone classifier of G-GBM is based on gradient-boosted decision trees. 
Such models are known to perform strongly in fraud detection settings due to their ability to capture non-linear relationships, handle heterogeneous feature types, and focus on difficult minority-class examples through iterative boosting. 
Second, our algorithm is equivalent to the typical supervised learning algorithm when considering only splits over the features of its head nodes. Our method has the added advantage of using further splits from neighbourhood observations. 
Therefore, our method G-GBM should perform at least as well as its classic supervised counterpart LightGBM.
Third, graph information is captured as an aggregation/concatenation of node and edge features of metapaths of a fixed length. 
This allows to assess which features led to the classification of a case as fraudulent and in what way. 
This is a very important consideration in the highly regulated insurance industry where the audit trail and explainability are of utmost importance. 
Finally, as the information at different distances is kept separate, G-GBM can distinguish among information coming from the different levels. 
This alleviates the problems of over-smoothing~\citep{Li_Han_Wu_2018}, which plague (H)GNNs when additional message passing layers are added. 

\section{Experimental Setup}
\label{sec:experiments}
We present an application to insurance fraud detection on two heterogeneous graphs.
The first is a real-world, proprietary insurance dataset, where we showcase the ``out-of-time'' prediction capabilities of G-GBM. 
The second is an open-source graph on healthcare provider (HCP) fraud. With the inclusion of open-source data, we aim for strong reproducibility of our results. 

The experiments are implemented in Python using the StellarGraph~\citep{StellarGraph} and PyTorch Geometric~\citep{Fey/Lenssen/2019, Fey/etal/2025} libraries. 
The full implementation is made available on GitHub ({\url{https://github.com/VerbekeLab/GBDT_Graphs}}).
All experiments were run on an Apple M3 pro chip with an 11-core CPU and 36 GB of memory. 
On top of the code, we publish the insurance company dataset on Kaggle({\url{https://www.kaggle.com/dsv/16002794}}), providing the community with a publicly available heterogeneous graph benchmark for fraud detection research. 

\subsection{Proprietary Dataset}
The real-world dataset contains information on companies in Belgium and concerns the relationships between two node types, companies and administrators, each having distinct node features~\citep{f_lix_vandervorst_bruno_deprez_tim_verdonck_wouter_verbeke_2026}. 
Under the Belgian Code of Companies and Associations, companies are legally required to register their directors and other statutory officers within their governance structure. 
These individuals may simultaneously hold mandates across one or multiple legal entities. 
Such overlaps in mandates reveal additional connections within corporate networks, that may be relevant for highlighting fraud.

Both administrators and companies can be associated with underwriting fraud.
Underwriting fraud can be defined as a category of insurance fraud occurring at the pre-contractual or renewal stages of the insurance lifecycle, encompassing intentional misrepresentation, omission, or fabrication of material information by the policyholder or applicant to obtain coverage under false pretences or to secure more favourable premium conditions. 
This form of fraud includes, but is not limited to, the concealment or distortion of risk-relevant information during the application process (application fraud), the manipulation of disclosed data to reduce premium levels (premium fraud), the non-disclosure of concurrent insurance contracts covering the same property and casualty risk, and the procurement of insurance coverage for non-existent or fictitious risks~\citep{viaeneUnderwriting}.

Since most of the nodes and edges are shared across the graphs over time, we distinguish between two time periods, to ensure the experimental setup resembles a realistic situation:
\begin{itemize}
    \item Information available at time $t_0$, i.e., a graph of companies and administrators captured at that point in time and the fraud labels collected historically up to that point in time.
    \item Information available at time $t_1$, i.e., an updated graph of companies and administrators (which can have newly added or deleted elements compared to $t_0$) and the fraud labels discovered between $t_0$ and $t_1$.
\end{itemize}

Table~\ref{tab:compa_Gt0Gt1} shows the basic characteristics of the graphs $\mathcal{G}_{t_0}$, $\mathcal{G}_{t_1}$. We also present their intersection graph \( \mathcal{G}_{t_0} \cap \mathcal{G}_{t_1}\) to highlight the persistent nodes and edges and determine the pace at which the graph evolves over the period.

\begin{table}[ht]
    \centering
    \caption{Real-world dataset: Two graphs \( \mathcal{G}_{t_0} \) and \( \mathcal{G}_{t_1} \) and their intersection \( \mathcal{G}_{t_0} \cap \mathcal{G}_{t_1}\)}
    \begin{tabular}{lrrr}
\toprule
                           &          $G_{t_0}$ &           $G_{t_1}$ &   $G_{t_0} \cap G_{t_1}$ \\
\midrule
 Number of Nodes           &     2.9e+6 &      3.1e+6 &                2.7e+6 \\
 - Company           &     1.5e+06 &      1.6e+6 &                1.3e+06 \\
 - Admin.     &     1.4e+06 &      1.6e+6 &                1.4e+06 \\
 Number of Edges           &     1.98e+6 &      2.2e+6 &                1.9e+6 \\
 - Company-Company       & 9.8e+4           & 10.8e+4           &            8.5e+4           \\
 - Company-Admin. &     1.9e+6 &      2.2e+6 &                1.8e+6  \\
 Average Degree            &     1.37      &      1.42     &                1.42     \\
 - Company                 &     1.40     &      1.49     &                1.51     \\
 - Admin.           &     1.34     &      1.34     &                1.33     \\
\bottomrule
\end{tabular}

    \label{tab:compa_Gt0Gt1}
\end{table}

We observe that company-to-company relationships are relatively scarce compared to company-to-administrator relationships but show a proportional increase in the latter time frame, as indicated by the increasing numbers from \( \mathcal{G}_{t_0} \) to \( \mathcal{G}_{t_1} \). The persistence of company-to-administrator edges in the intersection graph shows that those edges are relatively stable over the period.

The distribution of labels in each set is presented in Table~\ref{table:2x4}. The labels are collected in two time periods of equal length and contain positive and negative samples. 
\textit{Class balance} expresses how many of the investigated cases turned out to be actually fraud. 
\textit{Relative fraud} expresses the relative number of fraud cases compared to the full dataset of administrators and companies, respectively.

\begin{table}[ht]
\centering
\caption{Labels per node type for the proprietary insurance dataset.}
\begin{tabular}{ |c|c|c|c|c|c| } 
\hline
Time window & Node type & Labels & Fraud Labels & Class balance & Relative Fraud \\
\hline
$]t_{-1},t_0]$ & Admin & 7293 & 1627 & 22\% & 0.12\%  \\ 
\hline
$]t_0,t_1]$ & Admin & 5861 & 1584 & 27\% & 0.11\% \\ 
\hline
$]t_{-1},t_0]$ & Company  & 11150 & 5484 & 49\% & 0.37\% \\ 
\hline
$]t_0,t_1]$ & Company & 9295 & 5018 & 53\% & 0.34\% \\ 
\hline
\end{tabular}
\label{table:2x4}
\end{table}

\subsection{Healthcare Provider}
\label{subsubsec: data HCP}
We also test G-GBM on an open-source dataset to enhance reproducibility of our results. 
It contains possible healthcare provider fraud cases and is available on kaggle\footnote{\url{https://www.kaggle.com/datasets/rohitrox/healthcare-provider-fraud-detection-analysis}}. 
This dataset, unfortunately, does not contain ground truth labels of the test set, nor does it contain any time information on the label assignment. 
For the train-test split, we cluster the nodes using the Louvain method and assign entire clusters to either the training or test graph. This limits information leakage by ensuring that metapaths, and therefore the resulting feature vectors, remain within the corresponding train or test graph. 

Table~\ref{tab:hcp_G} shows the resulting graph characteristics. Table~\ref{table:labels_hcp} shows the label distribution, indicating a strong class imbalance. 
As said before, only providers are labelled. 

\begin{table}[ht]
    \centering
    \caption{Healthcare provider fraud}
    \begin{tabular}{lrr}
\toprule
                           &          $G_0$  & $G_1$ \\
\midrule
 Number of Nodes & 81,855 & 62,042  \\
 - Beneficiary & 79,255 & 59,261 \\
 - Provider & 2600 & 2781 \\
 Number of Edges & 205,386 & 146,205\\
 - Provider-Beneficiary & 205,386 & 146,205\\
 Average Degree & 5.02 & 4.71\\
 - Beneficiary & 2.59 & 2.47 \\
 - Provider & 78.99 & 52.57\\
\bottomrule
\end{tabular}
\label{tab:hcp_G}
\end{table}

\begin{table}[ht]
\centering
\caption{Labels for the healthcare provider dataset.}
\begin{tabular}{ |c|c|c|c| } 
\hline
Graph & Node type & Fraud Labels & Relative Fraud \\
\hline
$G_0$ & Provider & 278 & 11\%  \\ 
\hline
$G_1$ & Provider & 228 & 8\% \\ 
\hline
\end{tabular}
\label{table:labels_hcp}
\end{table}

Our method requires all nodes to have features, however, HCP dataset does not provide any features for the (labelled) provider nodes. 
Therefore, we construct four features based on the claims data for each provider. 
These are (1) the number of claims, the (2) average and (3) standard deviation of the amount paid, and (4) the number of beneficiaries. 
We use the provided features for the beneficiary nodes. 

\subsection{Baseline Methods}
Our proposed G-GBM method is tested against a couple of relevant baselines. 
First, given that G-GBM's backbone model is  LightGBM, we compare it to LightGBM~\citep{ke2017lightgbm} trained on only the node features. 
Second, node embeddings coming from metapath2vec~\citep{dong2017metapath2vec} are used to analyse the impact of the graph structure on the predictions. 
These node embeddings are once used on their own to make predictions and once in combination with the node features. 
The downstream classifier is again a LightGBM model. 
Third, we compare G-GBM to several established inductive HGNN baselines. These are HinSAGE, the StellarGraph~\citep{StellarGraph} implementation of GraphSAGE~\citep{hamilton2017inductive} for HINs, heterogeneous graph attention network~(HAT)~\citep{wang2019heterogeneous} and heterogeneous graph transformer~(HGT)~\citep{10.1145/3366423.3380027}.
These HGNNs incorporate the features and graph structure to come to a classification model. 

We note that metapath2vec is transductive, meaning that we need to construct the graph embedding for the full graph, containing both train and test data. 
We will only illustrate its use on the HCP dataset, since the train and test graphs are separated. 
For the proprietary dataset, most of the nodes are shared while labels and edges change, making metapath2vec impractical for this use-case. 

Table~\ref{tab:hyperparameters} summarises the choice of hyperparameters for the different methods. 
The default hyperparameters are used for LightGBM. 
For the HGNNs, we use configurations inspired by related fraud detection research~\citep{van2023catchm} and align the hidden dimensions and training settings across models where possible. 
The number of heads for HAN are also the default values. 
We halved the number of heads for HGT, since the calculations are more expensive to calculate per head. 

\begin{table}[]
\centering
\caption{Hyperparameter settings for all models.}
\label{tab:hyperparameters}
\begin{tabular}{llccccc}
\toprule
\textbf{Model} & \textbf{Hyperparameter} & \textbf{Value} \\
\midrule
\multirow{5}{*}{LightGBM}
  & Number of leaves       & 31    \\
  & Learning rate          & 0.05  \\
  & Feature fraction       & 0.9   \\
  & Boosting rounds        & 500   \\
  & Early stopping patience & 10   \\
\midrule
\multirow{4}{*}{Metapath2vec}
  & Embedding dimensions   & 64  \\
  & Number of walks        & 2   \\
  & Walk length            & 4   \\
  & Context window size    & 3   \\
\midrule
\multirow{6}{*}{HinSAGE}
  & Neighbourhood samples  & [10, 5]      \\
  & Layer sizes            & [32, 32]     \\
  & Dropout                & 0.5          \\
  & Learning rate          & 0.01 \\
  & Epochs                 & 10  \\
  & Early stopping patience & 5 \\
\midrule
\multirow{6}{*}{HAN}
  & Hidden size            & 64     \\
  & Attention heads        & 8      \\
  & Dropout                & 0.5    \\
  & Learning rate          & 0.001  \\
  & Weight decay           & 0.0001 \\
  & Epochs                 & 100    \\
\midrule
\multirow{7}{*}{HGT}
  & Hidden size            & 64     \\
  & Attention heads        & 4      \\
  & Number of layers       & 2      \\
  & Dropout                & 0.2    \\
  & Learning rate          & 0.001  \\
  & Weight decay           & 0.0001 \\
  & Epochs                 & 100    \\
  \midrule
\multirow{7}{*}{G-GBM}
  & Metapath length  & 2             \\
  & Path probability weighting & uniform       \\
  & Number of leaves           & 31            \\
  & Learning rate              & 0.05          \\
  & Feature fraction           & 0.9           \\
  & Boosting rounds            & 500           \\
  & Early stopping patience    & 10            \\
\bottomrule
\end{tabular}
\end{table}

\subsection{Evaluation Metrics}
Our experiments analyse both the predictive power and the computational efficiency of G-GBM. 
The predictive power is assessed using two metrics. The area under the ROC curve (AUC-ROC) is an established metric in assessing binary classification models. 
The ROC curve plots the true positive rate against the false positive rate. 
In the context of fraud detection, where the label distribution is highly imbalanced with much more legitimate cases, the AUC-ROC can give overly optimistic results due to the large number of (true) negatives, i.e., correctly labelled legitimate cases.
Therefore, the area under the precision-recall curve (AUC-PR) is preferred. 
The precision-recall curve plots the precision against the recall, with the recall the same as the true positive rate. 
Because precision is used instead of false positive rate, the AUC-PR is less sensitive to the large number of negative cases. 
Therefore, it is preferred to use the AUC-PR when evaluating fraud models. 

Next to predictive performance, we analyse the efficiency of our method. 
We report the training time of G-GBM and the baseline methods on the different datasets. 

Insights into feature importance are obtained using SHAP values~\citep{lundberg2017unified}. We also extract the (absolute) SHAP value, a common interpretation metric, to highlight the contribution of neighbourhood information.

\section{Results}
\label{sec:results}
\subsection{Company node prediction}
We compare our approach, G-GBM, which uses a local neighborhood of size 2, to the baseline GBM, which uses the features of the head node (i.e., the classical supervised learning framework without graph information). 
This comparison indicates the added value of local neighborhood information. 
Additionally, we benchmark our method against HinSAGE, HAN and HGT as challenger GNN approaches.

\begin{figure}
    \centering
    \includegraphics[width=\textwidth]{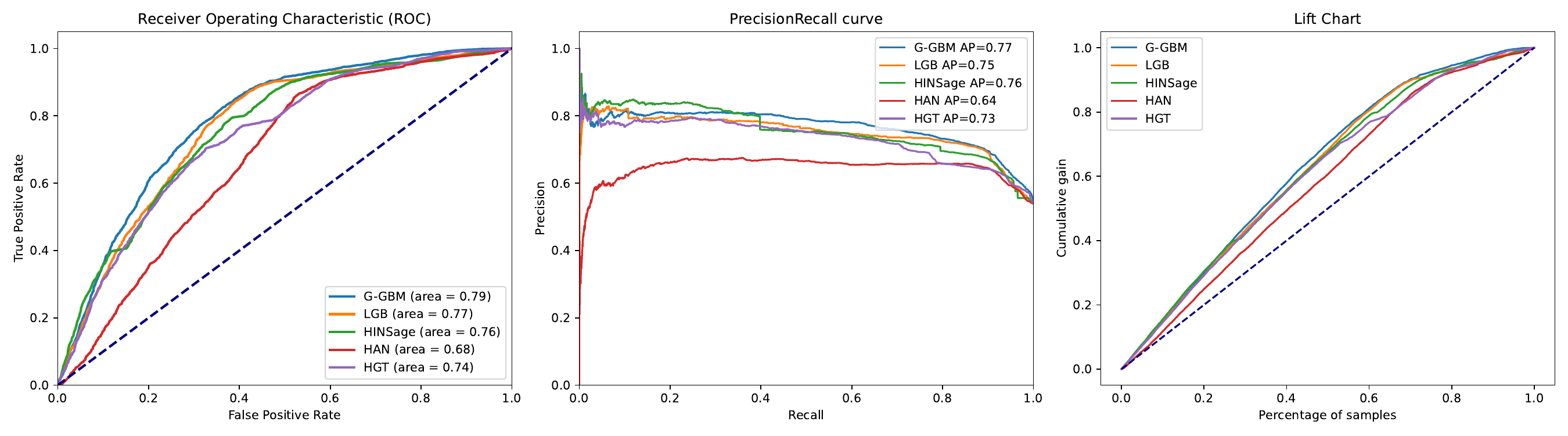}
    \caption{Performance measurements for the test set of company node labels.}
    \label{fig:comparison_company}
\end{figure}

The full results for the company node data are presented in Table~\ref{tab:res company}, with the ROC and PR curves presented in Figure~\ref{fig:comparison_company}.
We observe that gradient boosting outperforms all GNN models. 
However, our approach outperforms the baseline methods, with an increase in the ROC-PR of 0.02 over GBM, 0.08 over HinSAGE, 0.13 over GAT and 0.04 over HGT. 

\begin{table}[]
    \centering
        \caption{The experimental results for the company node prediction.}
    \begin{tabular}{l|cccc}
    \toprule
       Model&AUC-ROC&AUC-PR&Training time (s)\\
       \midrule
G-GBM&\textbf{0.79}&\textbf{0.77}&27.85\\
LGB&0.77&0.75&\textbf{0.46}\\
HinSAGE&0.76&0.76&4368.40\\
HAN&0.68&0.64&1905.47\\
HGT&0.74&0.73&2344.61\\
\bottomrule
    \end{tabular}
    \label{tab:res company}
\end{table}

\begin{figure}
    \centering
    \includegraphics[width=\textwidth]{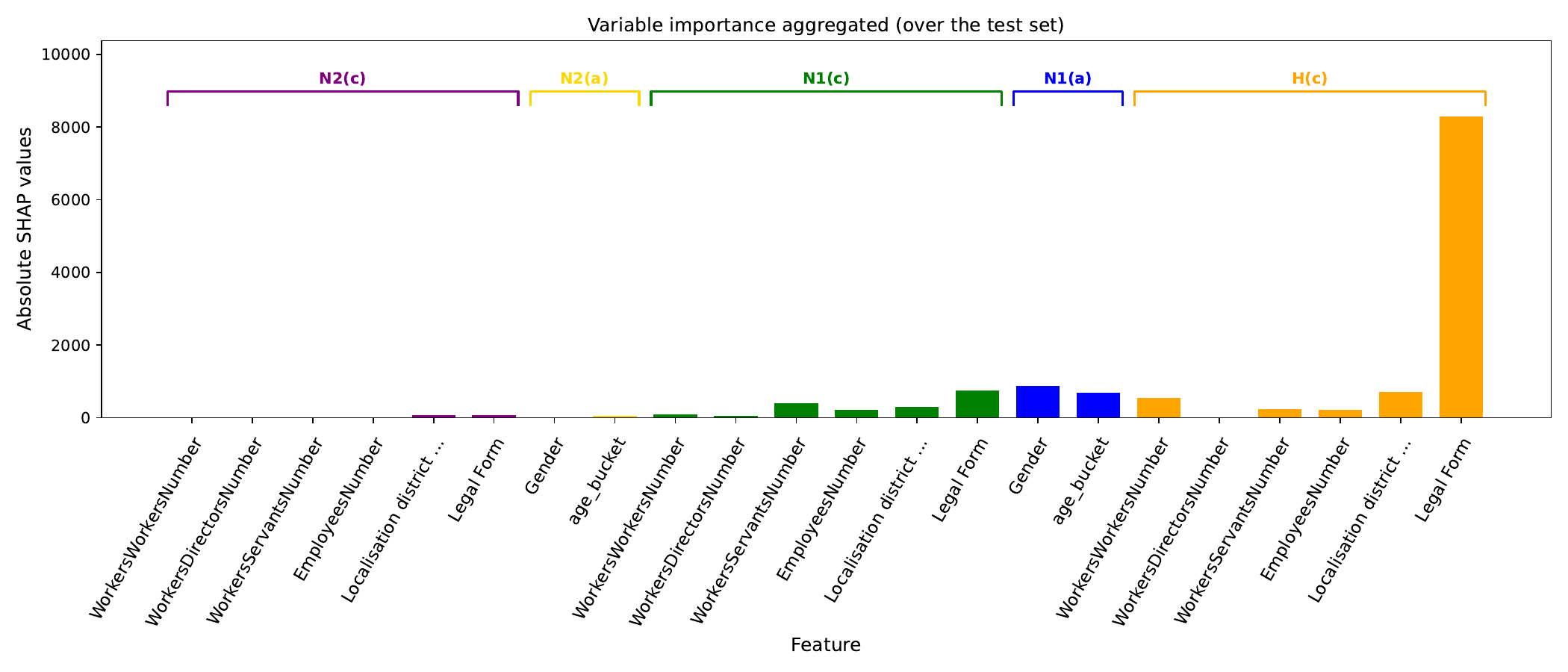}
    \caption{Variable importance in G-GBM: H is the head node, $N_q$ refers to the neighborhood of level $q$ relative to the head node, and (c) and (a) refer to the node type of the company and administrator, respectively.}
    \label{fig:shap_comp}
\end{figure}

The increased performance can be attributed to local characteristics, as illustrated by the SHAP values in Figure~\ref{fig:shap_comp}.
The SHAP values show that some features of the neighbouring administrator and company contain important information that are leveraged by the fraud detection model. 
The legal form of the company makes the greatest contribution to the G-GBM model prediction. 
This feature is also available for the baseline GBM model, which explains the models' relatively similar performance. 
We notice that nodes in the direct neighborhood contribute to the model performance, whereas the second neighborhood level is close to insignificant (which could motivate a reduction in the ego-net distance).

When considering training time in Table~\ref{tab:res company}, we see that only LightGBM is faster than G-GBM. 
This is to be expected, as G-GBM takes LightGBM as backbone and combines it with metapaths. 
The other HGNN models take much longer to train. 
HinSAGE has a shorter training time compared to the others, because of early stopping after 24 epochs. 

\subsection{Administrator node prediction}
\begin{figure}
    \centering
    \includegraphics[width=\textwidth]{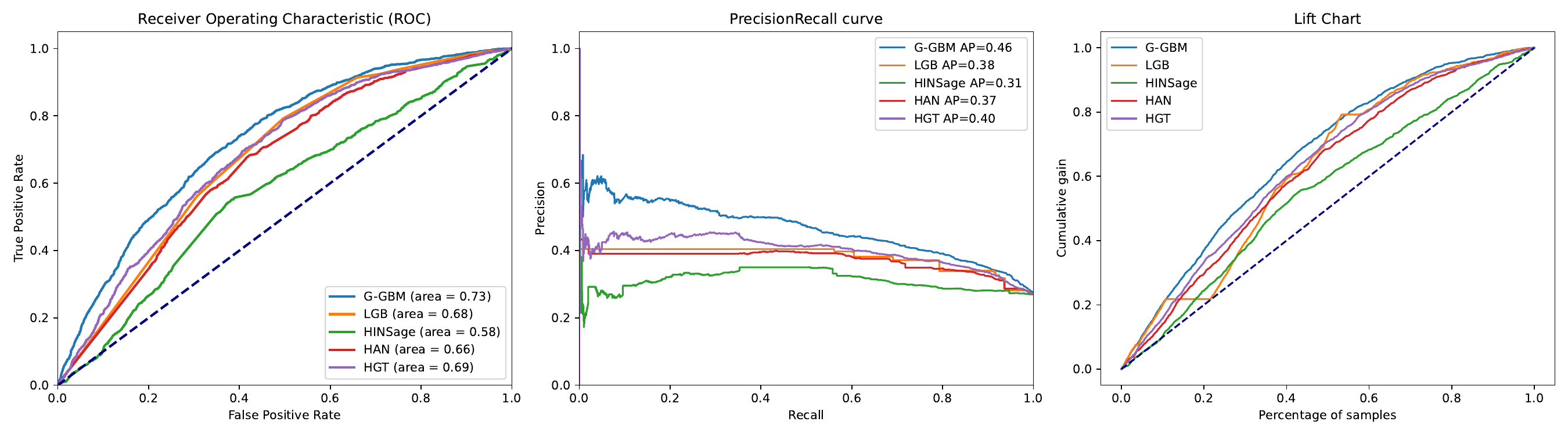}
    \caption{Performance measurement for the test set of administrator nodes (fraud vs. legitimate)}
    \label{fig:comparison_admin}
\end{figure}

This section evaluates the prediction of administrator labels using the same procedure. 
The results are given in Table~\ref{tab:res admin} and Figure~\ref{fig:comparison_admin}.
We observe that G-GBM achieves the highest AUC-ROC and AUC-PR among the evaluated models. 
The increase in performance of G-GBM is greater than in the previous experiment involving companies. 
This increase may be explained by the limited information available for head administrator nodes (only an age category and gender information), which limits discrimination power (hence, the piecewise linear appearance of the base GBM approach). 
Hence, in this case, the neighbourhood information is important.

\begin{table}[]
\centering
        \caption{The experimental results for the administrator node prediction.}
    \begin{tabular}{l|ccc}
    \toprule
       Model&AUC-ROC&AUC-PR&Training time (s) \\
       \midrule
G-GBM&0.73&0.46&14.64\\
LGB&0.68&0.38&0.03\\
HinSAGE&0.58&0.24&601.53\\
HAN&0.66&0.37&2290.97\\
HGT&0.69&0.41&2657.9\\
\bottomrule
    \end{tabular}
    \label{tab:res admin}
\end{table}

\begin{figure}
    \centering
    \includegraphics[width=\textwidth]{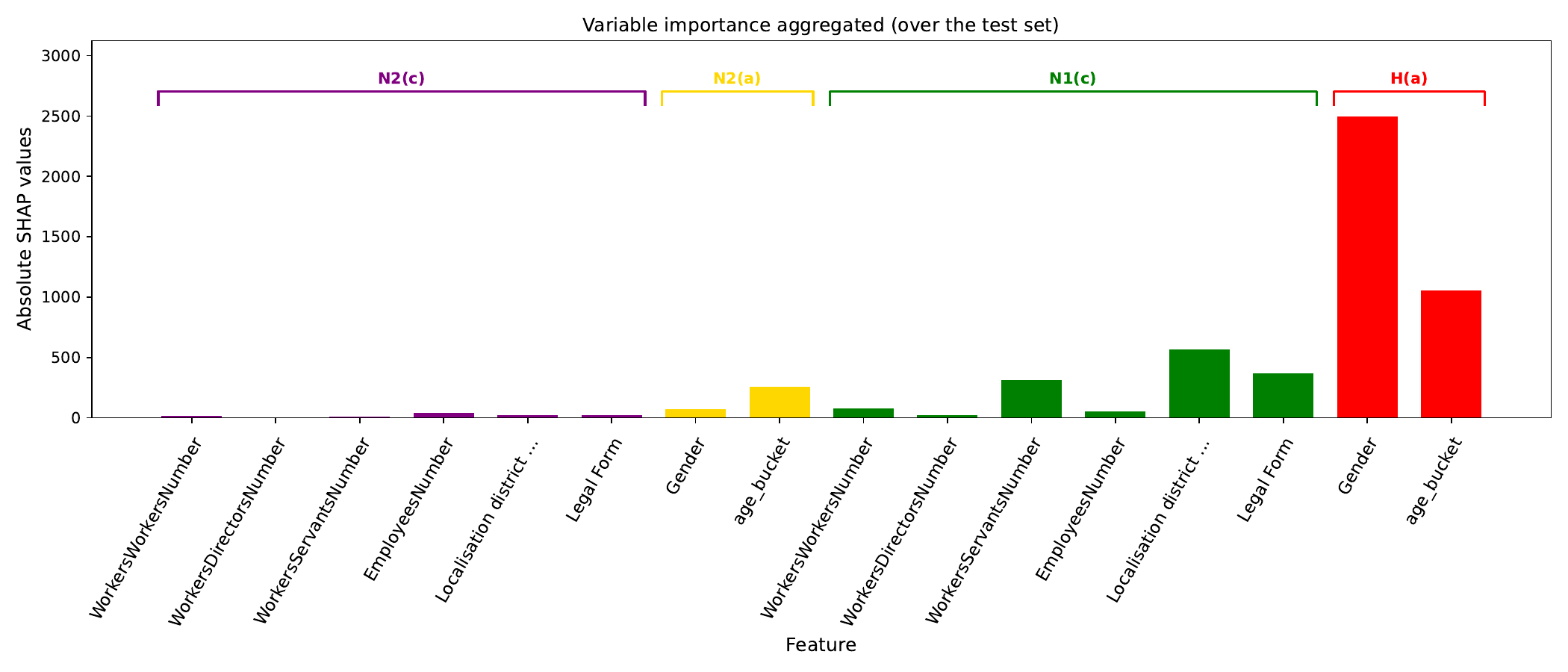}
    \caption{Variable importance in G-GBM: H is the head node, $N_q$ refers to the neighborhood of level $q$ relative to the head node, and (c) and (a) refer to the node type of the company and administrator, respectively.}
    \label{fig:shap-admin}
\end{figure}

The performance difference can be attributed to local characteristics, as shown by the SHAP-values in Figure~\ref{fig:shap-admin}.
Compared to the variable importance attribution in the company case, the contribution is more evenly spread between variables at different neighbourhood levels.
The greatest contribution is still coming from the head node, but features of nodes at distance two are now also important.

We also observe that HGT is the only other HGNN that outperforms the LightGBM baseline.
From the observation above, it seems that HinSAGE and HAN might suffer from oversmoothing, and are not able to capture the second-order neighbourhood structure well.

When considering training time in Table~\ref{tab:res admin}, we see that only LightGBM is faster than G-GBM. 
This is to be expected, as G-GBM takes LightGBM as backbone and combines it with metapaths. 
The other HGNN models take much longer to train. 

\subsection{Healthcare Provider}
For the HCP dataset, we only have labels for the providers. 
Because we are working with two separate graphs for training and testing, we can include experiments based on metapath2vec node embeddings, without the possibility of data leakage. 

\begin{figure}[H]
    \centering
    \includegraphics[width=\textwidth]{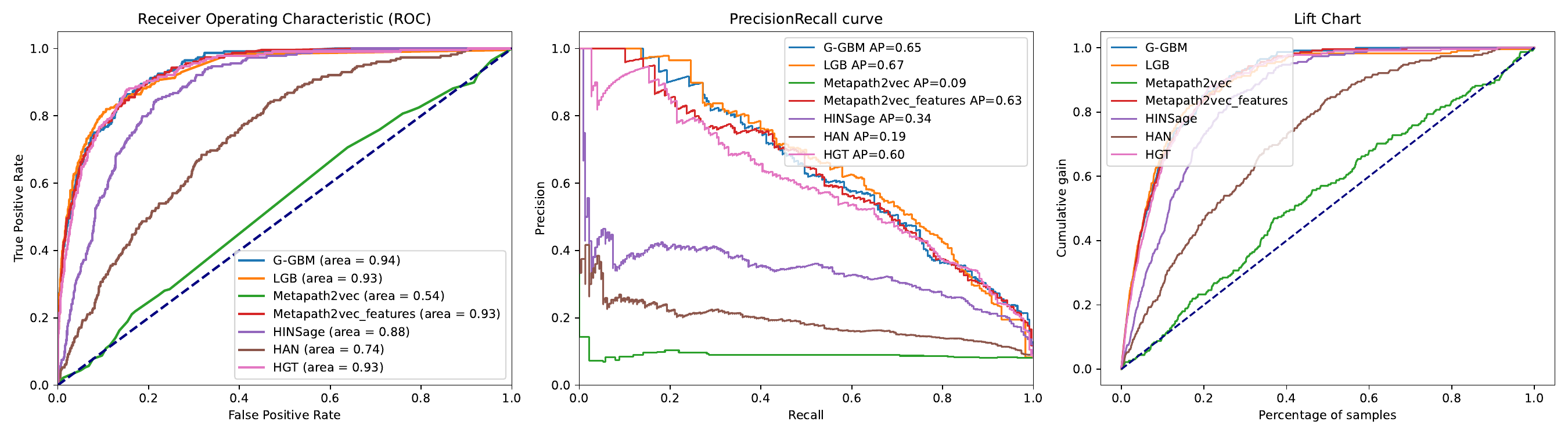}
    \caption{Performance measurement for the test set of HCP provider nodes.}
    \label{fig:comparison_hcp}
\end{figure}

The results are provided in Table~\ref{tab:res hcp} and Figure~\ref{fig:comparison_hcp}. 
The HCP dataset shows a setting in which the gains from graph-based neighbourhood information are less pronounced. 
In terms of AUC-ROC, we see only a slight improvement of our G-GBM model over the baseline LightGBM and metapath2vec with node features. 
These models all obtain very high performance, with an AUC-ROC above $90\%$. 
The AUC-PR on the other hand is slightly larger for LightGBM than for G-GBM. 
It seems that the inclusion of additional neighbourhood features that carry limited predictive signal may make it more difficult for G-GBM to identify fraudulent cases (see Figure~\ref{fig:shap_comp}). 
metapath2vec without additional features, on the other hand, is struggling to find any meaningful embedding, while HinSAGE and HAN obtains moderate performance, in terms of AUC-PR. 
This points to many false positives being present in the top predictions. 

\begin{table}[]
    \centering
        \caption{The experimental results for the healthcare provider fraud predictions.}
    \begin{tabular}{l|ccc}
    \toprule
       Model&AUC-ROC&AUC-PR&Training time (s)\\
       \midrule
G-GBM&0.94&0.65&366.61\\
LGB&0.93&0.67&0.08\\
Metapath2vec&0.54&0.09&3.67\\
Metapath2vec+features&0.93&0.63&3.63\\
HinSAGE&0.88&0.34&317.80\\
HAN&0.74&0.19&22.08\\
HGT&0.93&0.6&110.47\\
\bottomrule
    \end{tabular}
    \label{tab:res hcp}
\end{table}

\begin{figure}[H]
    \centering
    \includegraphics[width=\textwidth]{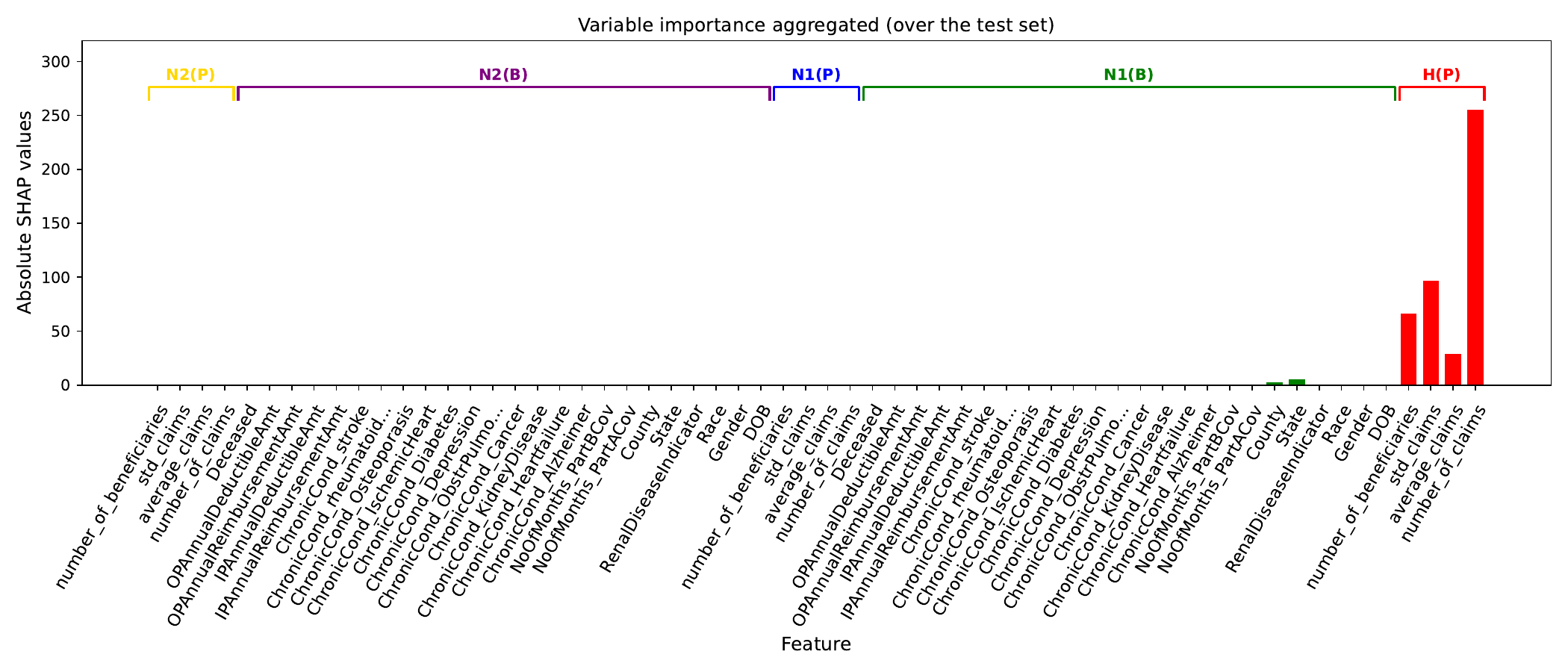}
    \caption{Variable importance in G-GBM: H is the head node, $N_q$ refers to the neighborhood of level $q$ relative to the head node, and (P) and (B) refer to the node type of the provider and beneficiary, respectively.}
    \label{fig:shap_hcp}
\end{figure}

The similarity between G-GBM, LightGBM and metapath2vec with features can be explained by looking at the variable importance plots in Figure~\ref{fig:shap_hcp}. 
We see that the predictions are primarily driven by the features of the head node, meaning that feature information from a node's neighbours does not add meaningful information. 
The fraud cases can already easily be detected using the four features we constructed in Section~\ref{subsubsec: data HCP}.

\begin{figure}
    \centering
    \includegraphics[width=0.95\linewidth]{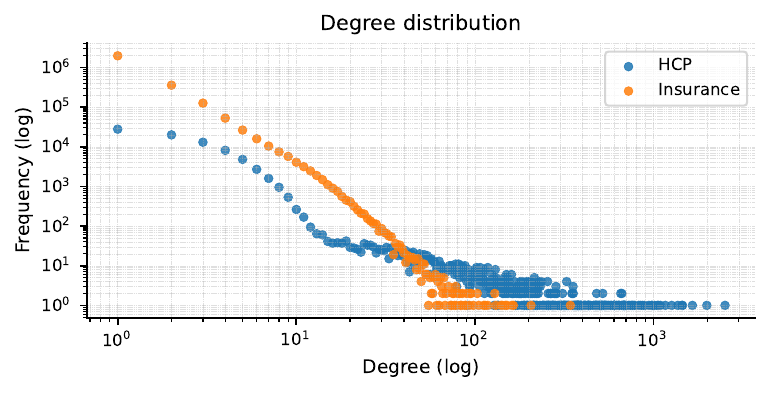}
    \caption{The degree distribution of the two datasets. }
    \label{fig:deg dist}
\end{figure}

Regarding computational performance, we see that G-GBM requires longer training time than the HGNN baselines on the HCP dataset. 
This is due to the topology of the HCP graph, and especially the degree distribution. 
The average degree of nodes in the graph is 5, while for the insurance dataset it is 1. 
Additionally, Figure~\ref{fig:deg dist} shows that, although the HCP graph is smaller than the insurance graph, it has more large hubs and the degree of these hubs is also larger than for the insurance dataset. 
This means that the number of possible metapaths explodes, making the computations less efficient. 

\section{Conclusion}
In this paper, we presented G-GBM, a novel method for insurance fraud detection on graphs based on boosted decision trees. 
These boosted decision trees are shown in the literature to be the dominant method for classification of tabular data.
We extended ensembles of gradient boosted trees to graph topologies by concatenating feature vectors alongside metapaths in an ego-net and use an elegant adaptation of case weights to construct trees.

We show that our method is consistent with supervised learning methods based only on ``intrinsic'' features, as it adds only new learnable information from the graph neighbourhood to the classifier, which is experimentally supported by the observed Pareto dominance of G-GBM over GBM in our experiments. 
The experimental results also show that our method outperforms popular inductive graph methods, in a supervised learning setup for fraud detection. 

Beyond predictive performance, the combination of metapaths with gradient boosting offers a set of practical advantages that are particularly valuable in the insurance industry. 
By encoding neighbourhood information as explicit concatenations of node and edge features along sampled paths, G-GBM retains the full interpretability of its LightGBM backbone. 
Unlike GNN-based approaches, which aggregate neighbourhood information into opaque embeddings, G-GBM produces predictions that can be decomposed using standard SHAP values and traced back to specific nodes and edges in the graph. 
This provides investigators with a clear audit trail linking a fraud prediction to concrete graph evidence. 
Combined with gradient boosting's native handling of class imbalance, categorical features, and missing data, this makes G-GBM a method that is not only competitive in terms of detection power, but also deployable in the highly regulated and audit-sensitive context of insurance fraud detection.

One downside of our method is the expansion of dimensions, which can become computationally impractical in the case of highly connected graphs, many node/edge types, high-dimensional features, or any combination thereof. 
On the other hand, the method does not require successive aggregation of features, a problem known to lead to over-smoothing in GNN architectures~\citep{Li_Han_Wu_2018}. 

Further research could elaborate on time-aware simple paths enable one to use previous records of fraud as a separate node while controlling for information leakage. 
Given the popularity of ``guilt by association'' methods in insurance fraud, we expect this representation to have the potential to reach another performance ceiling. 
Another promising avenue to improve our method could be the exploration of attention mechanisms, discarding uninformative paths. 
A possible simplistic extension could be the analysis of SHAP values grouped by metapath types, thereby discarding irrelevant metapaths and retraining the model. 
A more complex attention mechanism (e.g., \textit{there exists at least one path such that...}) would likely require more adaptation than using the case weights, thereby modifying the gradient-boosted model itself.

\section*{Acknowledgments}

We express our gratitude to Allianz for providing the data that enabled this research.

This work was supported by the Research Foundation – Flanders (FWO research projects 1SHEN24N and G015020N).

\bibliographystyle{agsm}
\bibliography{sn-bibliography}

\end{document}